%% file: main.tex
\documentclass[10pt,twocolumn,letterpaper]{article}

\usepackage{cvpr}              

\input{preamble}

\definecolor{cvprblue}{rgb}{0.21,0.49,0.74}
\usepackage[pagebackref,breaklinks,colorlinks,citecolor=cvprblue]{hyperref}

\def\paperID{31} 
\def\confName{EarthVision}
\def\confYear{2024}

\title{Cross-sensor super-resolution of irregularly sampled Sentinel-2 time series}

\author{Aimi Okabayashi\textsuperscript{1}, Nicolas Audebert\textsuperscript{2,3}, Simon Donike\textsuperscript{4}, Charlotte Pelletier\textsuperscript{1}\medskip\\
\textsuperscript{1} Université Bretagne Sud, IRISA, UMR CNRS 6074, F-56000 Vannes, France\\
\textsuperscript{2} Université Gustave Eiffel, ENSG, IGN, LASTIG, F-94160 Saint-Mandé, France\\
\textsuperscript{3} Conservatoire national des arts et métiers, CEDRIC, F-75141 Paris, France\\
\textsuperscript{4} Image Processing Laboratory (IPL), Universitat de València, València, Spain\\
{\tt\small nicolas.audebert@ign.fr, charlotte.pelletier@univ-ubs.fr}
}

\usepackage{censor}
\newcommand{\datasetname}{BreizhSR\xspace}
\newcommand{\urlcode}{\url{https://github.com/aimiokab/MISR-S2}\xspace}

\begin{document}
\maketitle
\input{sec/0_abstract}    
\input{sec/1_core_text}

\clearpage
{
    \small
    \bibliographystyle{ieeenat_fullname}
    \bibliography{main}
}


\end{document}

%% file: preamble.tex
\usepackage[dvipsnames]{xcolor}
\usepackage{xspace}
\usepackage{microtype}
\usepackage[detect-all]{siunitx}
\usepackage{tabularx}
\usepackage{booktabs}


%% file: sec/0_abstract.tex
\begin{abstract}

Satellite imaging generally presents a trade-off between the frequency of acquisitions and the spatial resolution of the images. Super-resolution is often advanced as a way to get the best of both worlds. In this work, we investigate multi-image super-resolution of satellite image time series, \ie how multiple images of the same area acquired at different dates can help reconstruct a higher resolution observation. In particular, we extend state-of-the-art deep single and multi-image super-resolution algorithms, such as SRDiff and HighRes-net, to deal with irregularly sampled Sentinel-2 time series. We introduce \datasetname, a new dataset for $4\times$ super-resolution of Sentinel-2 time series using very high-resolution SPOT-6 imagery of Brittany, a French region. We show that using multiple images significantly improves super-resolution performance, and that a well-designed temporal positional encoding allows us to perform super-resolution for different times of the series.
In addition, we observe a trade-off between spectral fidelity and perceptual quality of the reconstructed HR images, questioning future directions for super-resolution of Earth Observation data. The source code is available at \urlcode.
\end{abstract}

%% file: sec/1_core_text.tex
\section{Introduction}
\label{sec:intro}

Satellite imagery is one of the most powerful and effective tools to monitor and study the surfaces of the Earth, contributing to various applications such as weather forecasting, urban planning, or tracking natural disasters. However, its efficiency is constrained by trade-offs between spatial and temporal resolutions.
On the one hand, recent constellations can capture the same area with a high-revisit time, forming complex spatio-temporal data cubes coined as satellite image time series (SITS). A notable example is the two Sentinel-2 satellites that capture all land surfaces every five days at the equator at a 10-meter spatial resolution at best. The SITS data are commonly used to precisely monitor landscape dynamics, such as in land cover mapping or change detection, but their low to medium spatial resolution might be inadequate for some applications, such as urban mapping of buildings, roads, or sparse vegetation. On the other hand, very high-spatial resolution sensors, such as Pléiades Neo or WorldView, capture images at a metric or sub-metric resolution, useful for small object detection or individual building counting. For example, SPOT-6 acquires images at \SI{1.5}{\meter} spatial resolution. As well as often being commercial and thus necessitating the purchase of images, the lack of dense acquisitions poses challenges for continuous monitoring of dynamic landscapes, such as crops or forests.

\begin{figure}
    \centering
    \includegraphics[width=\linewidth]{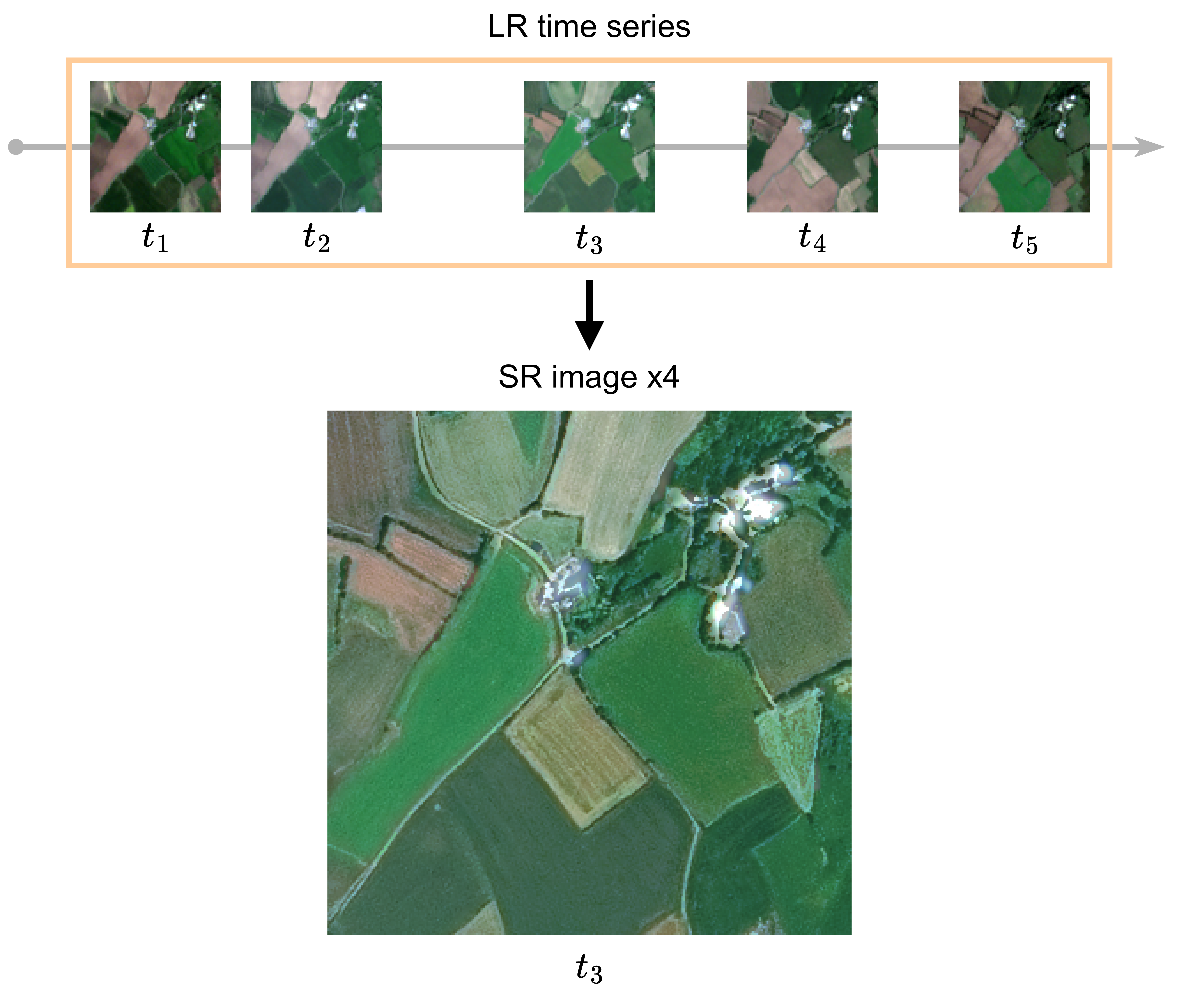}
    \caption{Multi-image super-resolution with an upscaling factor of~4. The irregular low-resolution (LR) satellite time series is used to predict a super-resolved (SR) image for a given acquisition date.}
    \label{fig:misr_challenges}
\end{figure}

One possible solution to make the most of high-temporal low-spatial data and low-temporal high-spatial data is \emph{super-resolution}, an image processing technique that aims to improve the spatial resolution of an image. In this context, the vanilla super-resolution ill-posed inverse problem consists of reconstructing a high-spatial resolution (HR) image from a low-spatial resolution (LR) image. This problem, 
coined as single-image super-resolution (SISR), has been tackled with an ever-growing corpus 
on deep learning, from CNN \cite{srcnn} to GAN \cite{srgan,rakotonirina2020esrgan+}, and now diffusion models \cite{sr3,srdiff}, yet with only preliminary applications to remote sensing \cite{khanna2023diffusionsat,xiao2023ediffsr}.

One challenge with super-resolution is that models are often trained on synthetic data, where (LR,HR) pairs are artificially created by downsampling an HR image \cite{timofte_ntire_2017,haut2019remote,lanaras_superresolution_2018}. In remote sensing, the synthetic LR image does not reflect what an actual LR sensor would capture in situ~\cite{wolters_zooming_2023}. Besides, this supposes that (LR,HR) were acquired simultaneously. The cross-sensor setting that uses different LR and HR sources is a way to overcome this issue, by training models robust to atmospheric effects, geometric distortions, and differences in spectral characteristics~\cite{michel2022sen2venmus}.

Additionally, access to multiple images of the same geographical area at different times can be beneficial to enhance the generation of an HR image by fusing information from multiple views~\cite{probavchallenge}. This approach, coined as multi-image super-resolution (MISR), applies super-resolution on a sequence of satellite images instead of a single image (see \cref{fig:misr_challenges}). MISR techniques usually extend SISR techniques with a dedicated module to learn from the sequence of LR images~\cite{kowaleczko2023real}. Among notable approaches, HighRes-net~\cite{highresnet} introduced a recursive fusion module and residual attention model (RAMS)~\cite{rams} proposed the use of residual channel attention to model the temporal relationships between LR images. However, these approaches consider regularly temporal sampled images while SITS are irregular and unaligned (not the same acquisition dates for different SITS) due to the multitude of satellite orbits and meteorological conditions. 

In this work, we study cross-sensor super-resolution techniques for SISR and MISR scenarios with diffusion models.
To exploit irregular and unaligned time series, we introduce a time-equivariant fusion module
to model the temporal relationships between LR images. Specifically, we improve on the light-weight temporal attention encoder (L-TAE), originally designed for SITS classification~\cite{ltae} and later on for SITS panoptic segmentation~\cite{ltaepanoptic}, to integrate the temporal dimension of SITS into MISR.
We evaluate our approaches on a new dataset, \datasetname, that consists of LR Sentinel-2 time series paired with HR SPOT-6 images acquired over the Brittany region located in northwest France. 

\noindent Our contributions are summarized as follows:
\begin{enumerate}
    \item We introduce \datasetname, a new dataset for real-world super-resolution of Sentinel-2 time series.
     \item We introduce a time-equivariant fusion module to handle irregularly sampled time series.
    \item We propose a super-resolution formulation that allows us to generate an HR image at any specified date within the time frame of the LR time series.
\end{enumerate}

\section{Related Work}
\label{sec:related}

Super-resolution (SR) techniques have witnessed a shift from traditional interpolation-based methods to data-driven deep-learning approaches in recent years. Traditional methods, such as bicubic interpolation or Lanczos resampling, yielded limited improvements and failed to capture the intricate details and high-frequency information in the super-resolved images \cite{Yang2019}. In this section, we present the state-of-the-art methods for SISR and MISR. We will focus on deep learning approaches since they outperform traditional methods~\cite{srsurvey}. 

\textbf{Single-image super-resolution} (SISR) techniques include standard neural networks, attention-based models, generative adversarial networks (GANs), and diffusion models.
Pioneer deep super-resolution based on convolutional neural networks \cite{srcnn} quickly evolved to integrate architecture improvements such as residual learning~\cite{vdsr,edsr} and channel attention~\cite{rcan}. 
State-of-the-art models are now based on generative models, such as GANs. For example, SRGAN~\cite{srgan} combines a perceptual loss with an adversarial loss, and is one the first techniques reaching a $4\times$ upscaling factor. It was refined in ESRGAN \cite{esrgan} using the residual-in-residual dense block (RRDB) generator and an improved GAN formulation. The latest models rely on diffusion models, defined by denoising diffusion probabilistic models (DDPM)~\cite{ddpm}. In brief, they gradually inject random noise into the input image and then learn to reverse the diffusion process to generate new samples from the noise. State-of-the-art diffusion models such as \cite{sr3} have shown their ability to model complex image distributions and to outperform both CNN and GAN-based approaches. Newer works, such as SRDiff~\cite{srdiff}, experiment with conditioning the diffusion model using a classical super-resolution model, \eg RRDB.

In remote sensing, super-resolution has been proposed as a way to extract HR information from LR imagery \cite{wang2022comprehensive}. 
Similarly to classical computer vision, CNN~\cite{lanaras_superresolution_2018,haut2019remote} or GAN~\cite{zhang2020approach} can be trained to super-resolve downsampled remote sensing images.
The drawback of these approaches is that the LR data is synthetic, and simulated by degrading real images. To close the gap with real data, many works have proposed ways to improve the synthetic data by
using better spectral models \cite{tarasiewicz_semisimulated_2022a} or simply 
different sensors for the LR and HR images \cite{galar2020super,zhu_superresolving_2020,wolters_zooming_2023}. This setup, called \emph{cross-sensor super-resolution} is more realistic and promising to apply SISR models to real-world data, but poses other problems. It requires co-registered LR and HR acquisitions of the same area, at similar dates to avoid changes. 
Several datasets have been introduced, focusing mainly on Sentinel-2~\cite{michel2022sen2venmus}, whose aliasing and inter-band shift characteristics have shown to be particularly well-suited for super-resolution \cite{nguyen_role_2023}.

\textbf{Multi-image super-resolution} (MISR) extends SISR by exploiting multiple LR images to generate an HR image. The idea is that each LR image in the sequence contains a different portion of the HR information. Most approaches fuse the information from various LR perspectives captured within a limited time frame to produce an HR image that depicts the landscape during this period. Formally, we aim at generating a super-resolved image $\widehat{HR}$ applying some function $h$ parameterized by $\theta$ to the set of LR images $(LR_1, ..., LR_T)$, with $T$ the number of images in the sequence, \ie
\begin{equation}
    \widehat{HR} = h_\theta(LR_1,..., LR_T)
    \label{eq:formulation_misr1}
\end{equation}

Most MISR approaches are designed for video processing \cite{caballero_realtime_2017,haris_recurrent_2019}, which typically assume a fixed time interval between frames and a high frame rate. These hypotheses are not reasonable for satellite imagery perturbed by the presence of clouds or saturated pixels.
Using the information from multiple images should help generate images that are more robust to these perturbations.
MISR datasets in remote sensing are less common. The most known is the Proba-V challenge, a competition organized by the European Space Agency~\cite{probavchallenge}, that sparked a wave of interest in MISR for remote sensing. However, Proba-V uses a fixed number of images for each series, with a constant time gap between two images. Closest to our work, MISR MuS2~\cite{kowaleczko2023real} and WorldStrat~\cite{cornebise_open_2022} datasets use Sentinel-2 SITS as LR data, and WorldView-2 or SPOT-6 images, respectively, for the HR reference. 

With these new datasets, novel architectures have been proposed to incorporate the temporal dimension in super-resolution models. Most models are built upon an encoder-decoder structure, as in SISR, but with a modified encoder to incorporate multiple images as inputs.
One of the seminal works in MISR for remote sensing is HighRes-net~\cite{highresnet}, a network composed of an encoder, a recursive fusion network, and a decoder. More precisely, a reference image ($\text{ref}$) defined as the median of the LR image time series is computed and concatenated with each LR image $LR_i$ ($1\le i\le T$). A hidden state is then computed for each concatenation $[LR_i, \text{ref}]$. The fusion network then fuses the hidden states recursively, by halving by two the number of LR states at each fusion step. The decoder reconstructs the HR image from the fused state with a deconvolution layer followed by a final convolutional layer.
The RAMS network~\cite{rams} uses a combination of residual feature attention to reduce the temporal dimension of the input and channel attention to weigh the different images. 3DRRDB \cite{ibrahim_3drrdb_2022} extends RRDB to multiple images by replacing the 2D convolutional layers by 3D convolutions.
Finally, TR-MISR \cite{trmisr} adds learnable channel attention to the HighRes-net encoder. A limitation of these methods is that they assume a fixed number of input images and a constant acquisition rate. While this holds for Proba-V, this is a problem for many practical applications with Sentinel-2 imagery, that we will address in this work.

\section{Proposed methods}
\label{sec:methods}

\subsection{From single- to multi-image super resolution}

Diffusion models are already very popular in SISR applied to synthetic data, such as the two DDPM models: SR3~\cite{sr3} and SRDiff~\cite{srdiff}. In this work, we propose to study SRDiff~\cite{srdiff} on a cross-sensor scenario.
SRDiff uses a U-Net~\cite{unet} network in the reverse process to estimate the noise, which is fed with the noisy image conditioned (\ie concatenated) by LR information. While a simple option is to condition the U-Net model directly with the LR image~\cite{sr3}, SRDiff conditions it with an LR encoder to allow the extraction of valuable features from the LR image. The LR encoder used in SRDiff is RRDB, the generator of ESRGAN~\cite{esrgan}, which combines multi-level residual networks and dense connections.

In the case of SRDiff, the diffusion model acts like a decoder: the model is conditioned by the output of the backbone model (\eg RRDB) to predict the residual between the upsampled LR image and the HR image. Therefore, extending SRDiff to the multiple image setting is a simple matter of switching RDDB to a MISR backbone. A solution is to use an existing MISR backbone such as HighRes-net or to extend RRDB to the MISR case. This is possible by individually encoding each image in the sequence with the RRDB encoder, then proceeding to fuse the hidden states with a MISR fusion module, such as the recursive module of HighRes-net or our time-equivariant module, and finally using the decoder to obtain the super-resolved image. We will use both replacement and extension strategies in the rest of this work.

\subsection{Time-Equivariant Super-Resolution}
\label{subsec:time_sr}

The MISR formulation provided by \cref{eq:formulation_misr1} assumes regular acquisitions within a small time frame. It sets a lot of constraints on the acquisition, as satellite image acquisition is limited by several factors, from the revisit rate to the presence of clouds or other artifacts, making some (parts of) images unusable. Besides, SITS can present temporal and spatial decorrelations, \ie the image content can change between two acquisitions due to gradual (\eg vegetation growth) and abrupt changes (\eg harvests and urban expansion).
A caveat of the MISR approaches is that they assume that the time gap between two images is constant or all images are equal, \ie all images in the sequence contribute the same to the reconstruction of the HR output. However, as we have seen before, none of these assumptions hold. 
Learning to weigh each image in the LR time series should enable super-resolution models to take account of the perturbations affecting the LR image and to decide which LR images should contribute most to the generation of the HR image.

To deal with this problem, we propose a time-dependent MISR approach that relies on the acquisition dates of LR and HR images. The idea is to acknowledge the time differences within the LR time series, but also their time difference with the HR acquisition. Therefore, the generated solution will consider information from the entire set of images, assigning different weights to each image. Based on a set of $T$ images with acquisitions dates at $t_1, ..., t_T$, and a given time $t_\text{HR}$ within the time span, we generate the HR image at a time $t_\text{HR}$ through a deep neural network $h$ of parameters $\theta$:
\begin{equation}
    \widehat{HR}(t_\text{HR}) = h_\theta\big((LR_1,t_1),..., (LR_T, t_T), t_\text{HR}\big)
    \label{eq:formulation_misr2}
\end{equation}
This formulation enables the generation of an image for any specified time within the given time frame.

We use the structure of MISR techniques and propose as a fusion module the lightweight temporal attention encoder (L-TAE)~\cite{ltae}, developed originally for crop-type identification. 
It is a simplified version of the multi-head self-attention network~\cite{attentionisallyouneed} designed for fast computation and efficiency purposes. The self-attention allows each image in the sequence to contribute differently to the model output. This helps deal with changes in the time series: images that are not in agreement with the HR image will receive less attention. In practice, we use the 2D version of L-TAE adopted for panoptic segmentation~\cite{ltaepanoptic}. It applies a shared L-TAE with multiple heads applied independently at each pixel. The idea is to enable individual weighting of the pixels within the patch to account for the diverse landscapes and the presence of local perturbations such as clouds and their shadows.
For simplification, we will use L-TAE to designate L-TAE 2D in the following.

To include the temporal information, \citet{ltae} originally define the positional encoding used in the Transformer layer as the number of days elapsed since the beginning of the sequence, \ie for $k=1,\ldots,T$:

\begin{equation}
    p^{(k)} = \left[\sin\left(\text{day}(t_k)/\tau^{\frac{i}{c_e/H}}\right)\right]_{i=1}^{c_e/H}
    \label{eq:tem_pos_absolute}
\end{equation}
where $\text{day}(t_k)$ denotes the number of days between $t_k$ and the beginning of the sequence, $\tau = 1000$ is the characteristic time scale, $c_e$ is the embedding dimension, and $H$ the number of attention heads.
This positional encoding is \emph{absolute} regarding the beginning of the time series, and therefore might not be well-suited to changes in the support of the time series. For example, removing the first image of the sequence changes the temporal encoding.

In contrast, we redefine the positional encoding to use the acquisition date of the target HR image as a reference point. In practice, we replace $\text{day}(t_k)$ in \cref{eq:tem_pos_absolute} with the difference between the date of the HR reference and the date of each LR image in the sequence. Formally, for $k=1,\ldots,T$, the positional embedding vector $p^{(k)}$ of dimension $c_e/H$ is given by:
\begin{equation}
    p^{(k)}=\left[\sin \left( (t_k-t_\text{HR})/ \tau^{\frac{i}{c_e/H}}\right)\right]_{i=1}^{c_e/H}
    \label{eq:tem_pos_encoding}
\end{equation}
Changing the temporal support of the time series, \ie removing or adding an image, does not alter the encoding of the other images.
At training time, the model $h_\theta$ is trained using the actual $t_\text{HR}$ date from the reference HR image. At inference, the reference time can be set to any time between $t_1$ and $t_T$.
This means that the same sequence of images $(LR_1, t_1), \dots, (LR_T, t_T)$ can be super-resolved at various reference times and produce different images $\text{HR}_{t}$.

\section{Experimental setup}

The source code and the dataset are available at \urlcode.

\begin{table*}
    \small
    \centering
    \setlength{\tabcolsep}{3pt}
    \begin{tabularx}{\textwidth}{rlXllXXXX}
        \toprule
        Dataset & Setting & Area & Dates (LR) & Dates (HR) & LR sensor & HR sensor & \# LR img\\
        \midrule
        SEN2VEN$\mu$S \cite{michel2022sen2venmus} & SISR & \SI{218000}{\kilo\meter\squared} & \small 12/2017 -- 10/2020 & Same day & S2 \SIrange[range-units=single,range-phrase=\,--\,]{10}{20}{\meter} & VEN$\mu$S \SI{5}{\meter} & 1\\
        Satlas \cite{wolters_zooming_2023} & SISR & \SI{123000}{\kilo\meter\squared} & $\pm$ 30 days & 2019--2020 & S2 \SI{10}{\meter} & NAIP \SI{0.6}{\meter} & 1\\        
        Proba-V \cite{probavchallenge} & MISR & \SI{109000}{\kilo\meter\squared} & $\pm$ 15 days & ? & Proba-V \SI{300}{\meter} & Proba-V \SI{100}{\meter} & $\geq 9$\\        
        WorldStrat \cite{cornebise_open_2022} & MISR & \SI{10000}{\kilo\meter\squared} & $\pm 40$ days & 2017 -- 2019 & S2 \SI{10}{\meter} & SPOT-6 \SI{1.5}{\meter} & 16\\
        MuS2 \cite{kowaleczko2023real} & MISR & \SI{3200}{\kilo\meter\squared} & \small  04/2019 -- 03/2021 & \small  09/2010--03/2015 & S2 \SI{10}{\meter} & WV-2 \SI{0.4}{\meter} & 14--15\\        
        \midrule
        \textbf{BreizhSR} & MISR & \SI{35000}{\kilo\meter\squared} & $\pm$ 60 days & \small  05/2018 -- 09/2018 & S2 \SI{10}{\meter} & SPOT-6 \SI{1.5}{\meter} & 8--26\\
        \bottomrule
    \end{tabularx}
    \caption{Comparison of dataset characteristics for SISR/MISR of remote sensing images.}
    \label{tab:dataset_char}
\end{table*}

\subsection{The \datasetname dataset}

\begin{figure}
    \centering
    \includegraphics[width=\linewidth]{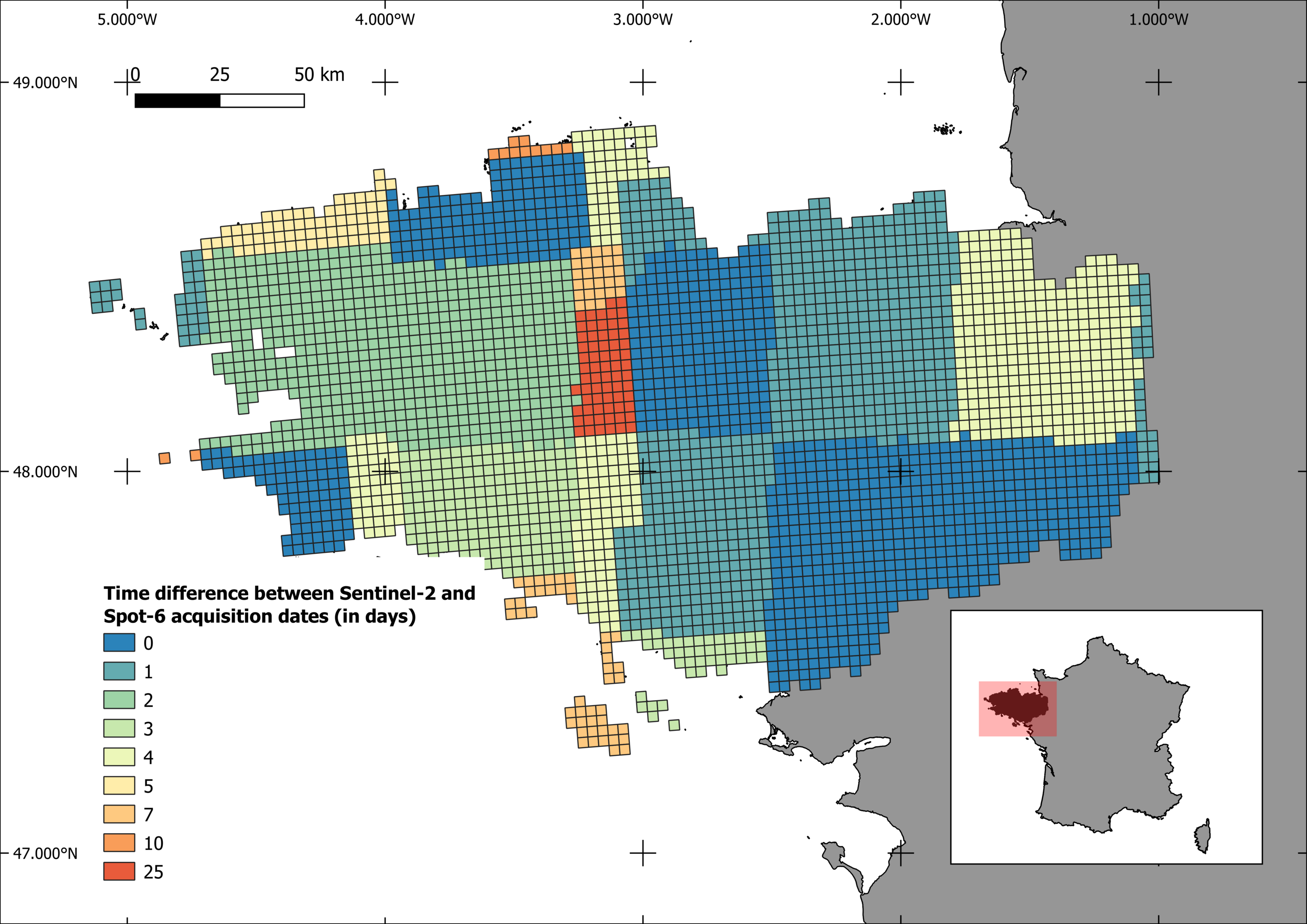}
    \caption{Study area with the minimal time difference of Sentinel-2 and SPOT-6 acquisitions.}
    \label{fig:study_area}
\end{figure}

To evaluate the methods proposed in \cref{sec:methods}, we propose \datasetname, a cross-sensor MISR dataset. The study area is the region of Brittany (\textit{Breizh} in the local language), located on the northwestern coast of France with an oceanic climate (see \cref{fig:study_area}). It covers about \SI{35000}{\kilo\meter\squared} with mostly agricultural areas (about 80~\%). 
The dataset comprises cross-sensor satellite data, with low-spatial but high-temporal resolution image time series sourced from Sentinel-2 satellites and the corresponding low-temporal high-spatial resolution images provided by the SPOT-6 satellite.

\textbf{Sentinel-2} constellation has twin satellites launched by the European Space Agency (ESA) in 2015 and 2017 that cover all Earth's surfaces every five days at the equator. Level-2A images of the \datasetname dataset are gathered \textit{via} the THEIA platform, which employs the MAJA pre-processing algorithm to obtain atmospherically corrected ground reflectance~\cite{hagolle2015multi}. 
To match the SPOT-6 spectral characteristics, only RGB bands at a 10-meter spatial resolution (B4, B3, and B2) are used in the analysis. The images were collected for the nine tiles covering the Brittany region from the 1st of April 2018 to the 31st of August 2018, filtering images with a cloud cover under 5~\%. Since the SPOT-6 data was acquired in the summer of 2018, the Sentinel-2 time period was chosen to include images from before and after the SPOT-6 acquisitions while staying in a range of similar seasonal and climate conditions.

\textbf{SPOT-6} is an HR optical Earth-imaging satellite system operated by the French national space agency (CNES) and launched in 2012. It can capture daily about 6~million \SI{}{\kilo\meter\squared} at a spatial resolution of up to 1.5~meters. SPOT-6 is used for commercial purposes and thus operates mainly on customer demand.
In this work, we focus on pan-sharpened RGB images with a spatial resolution of 1.5 meters. The used products are radiometrically and geometrically corrected L2A images with an 8-bit value range.
They consist of 3\,352 RGB image tiles of size 3000 $\times$ 3000 px, acquired between the 19th of April and the 3rd of August 2018. 

\textbf{Temporal discrepancy}
Given the high temporal resolution of the Sentinel-2 data, the vast majority of tiles show a low temporal difference from the SPOT-6 image to the temporally closest Sentinel-2 image, with over 78~\% of the SPOT-6 images having a corresponding Sentinel-2 image with a difference of 10 days or closer, as can be seen in \cref{fig:study_area}. The median number of Sentinel-2 images in the time series is 8, and the maximum is 17.

\textbf{\datasetname uniqueness} Compared to existing datasets, \datasetname exhibits several unique characteristics as highlighted in \cref{tab:dataset_char}. It complements existing datasets with wider coverage, smaller than Proba-V but at a significantly higher spatial resolution, on a limited geographical region but without focusing on specific areas such as WorldStart~\cite{cornebise_open_2022} or MuS2~\cite{kowaleczko2023real} that cover only urban areas. In addition, WorldStrat performs no filtering of cloudy images, reducing the number of effectively usable images in the series. Finally, \datasetname uses LR and HR acquisitions that are close in time (less than two months apart, with a median time difference below $10$ days) to mitigate seasonal effects. This strategy differs from the MuS2 dataset, which depends on open WorldView-2 HR images separated by several years from the LR series.

\subsection{Pre-processing}

We perform super-resolution with an upscaling factor of 4 (see \cref{fig:misr_challenges}). To reach this factor, the SPOT-6 images were downsampled through bicubic interpolation from \SI{1.5}{\meter} to \SI{2.5}{\meter}. Sentinel-2 and SPOT-6 images were respectively cropped into $3\times74\times74$ px and $3\times296\times296$ px.
In total, there are 52\,255 S2 time series and SPOT-6 image pairs.

The images stem from two different sensors, Sentinel-2 and SPOT-6, for which the value range and distribution are different. This is not only caused by differences in weather and radiation conditions at the time of acquisition but also by inner differences in terms of spectral bands, atmospheric correction, radiometric resolution and data processing. To mitigate the spectral discrepancy and help the models' convergence, we first min-max normalise both data sources independently, with statistics computed on the whole training set. In practice, 
we compute 2~\% and 98~\% percentile as an estimation of minimum and maximum values of Sentinel-2 data to take into account the presence of outliers due to artifacts such as clouds and their shadows. 
Then, for each pair of images $(LR, HR)$ in the SISR setting or $([LR_i]_{i=1}^T, HR)$ in the MISR setting, we perform a histogram matching of the HR image from SPOT-6 towards the distribution of the temporally closest Sentinel-2 LR image.

To evaluate the performance of the proposed methods, the dataset is split into training and test sets using a random block sampling strategy. Blocks of 4 $\times$ 4 SPOT-6 tiles (\ie 10~$\times$~10~km\textsuperscript{2}) are formed to ensure independence between training and testing sets. Out of 52\,255 image pairs, 37\,234 compose the training set (70~\%), including 10\% reserved for validation, and 15\,021 compose the test set (30~\%).

\begin{table*}
\small
\centering
  \begin{tabular}{@{}lccccccc@{}}
    \toprule
    Method & SR setting & $\downarrow$MAE  & $\downarrow$Shift-MAE & $\downarrow$RMSE    & $\downarrow$LPIPS & $\uparrow$PSNR  & $\uparrow$SSIM \\
    \midrule
    RRDB              & SISR       & 19.27          & 18.18          & 28.96          & 0.60          & 20.03          & 0.46          \\ 
    SRDiff bicubic upsample             & SISR       &24.19 & 22.87     & 35.54 & 0.33  & 18.17 & 0.37 \\
    SRDiff RRDB                         & SISR       &23.06 & 21.82     & 34.11 & \textbf{0.32}  & 18.52 & 0.39 \\
    HighRes-net recursive fusion & MISR &  23.33          & 22.67          & 33.89          & 0.52          & 18.50          & 0.42          \\
    RRDB L-TAE        & MISR       & 18.95          & 18.10          & \textbf{28.45}          & 0.59          & 18.95          & 0.46          \\
    HighRes-net L-TAE  & MISR & \textbf{18.79} & \textbf{18.00}         & 30.98 & 0.49          & \textbf{20.00}          & \textbf{0.45}         \\
    SRDiff HighRes-net L-TAE            & MISR      &24.62      &23.67           &35.75       &0.34       &17.90       & 0.36      \\
    \bottomrule
  \end{tabular}
  \caption{Quantitative results for a $4\times$ upscaling factor. Bold values highlight the best results.}
  \label{tab:encoders}
\end{table*}

\subsection{Implementation and architectures}

For the SISR setting, we assess the performance of diffusion models and conditioning by comparing two existing models--SRDiff conditioned with the LR image upsampled through bicubic interpolation and the output of the RRDB pre-trained model-- with the original RRDB super-resolution model.
For the MISR setting, we assess the benefits of the fusion module based on temporal attention mechanisms (\ie L-TAE) by experimenting with three models: (i)~HighRes-net,
(ii)~RRDB,
and (iii)~SRDiff conditioned by HighRes-net with L-TAE as the fusion module. We compare the results to the original HighRes-net using the recursive fusion module.

All networks are trained using the $L_1$ loss and Adam optimizer on an NVIDIA A100 80GB. HighRes-net recursive fusion uses two encoding layers and is trained for 300k training steps with a batch size of 32, a learning rate of $6e-4$ and a $0.7$ decay every 50k. The same architecture is used to train HighRes-net L-TAE. 
In both SISR and MISR settings, RRDB has 8 blocks and is trained with a learning rate of $2e-4$ and a batch size of 10.
SRDiff uses 500 diffusion steps and a batch size of 64. We train for 400k steps in SISR and 325k in MISR (SRDiff HighRes-net L-TAE).
In MISR experiments, the length of the time series is set to 8.

\subsection{Evaluation metrics}

To assess the performance of the SISR and MISR methods, we compute widely used quality measures averaged over all the test images. In particular, we use three reconstruction-based metrics--mean absolute error (MAE), shift-MAE, and root mean square error (RMSE)-- and three perception-based metrics--peak signal-to-noise ratio (PNSR), learned perceptual image patch similarity (LPIPS)~\cite{lpips}, and structural similarity index measure (SSIM)~\cite{ssim}.

\textbf{Shift-MAE} Slight misalignments between satellite images can occur, especially when using two different sources. To compensate for the error induced by those pixel shifts, we consider the Shift-MAE metric, inspired by the scoring method used in the Proba-V challenge~\cite{probavchallenge}. It consists in computing the MAE on sub-images shifted, and keeping the best value: $  \operatorname{Shift-MAE} = \min_{u, v \in\{0, \ldots, \delta\}}\operatorname{MAE}(\mathrm{HR}_{u, v},\operatorname{SR_{center}})$, where $\delta$ is the maximum shift allowed (in this work, $\delta =6$, \ie 15 meters) and  $\operatorname{SR_{center}}$ is the super-resolution image cropped to avoid the border effect and $\mathrm{HR}_{u,v}$ the sub-images with its upper left corner at coordinates $(u,v)$.

\section{Results and discussion}
\label{sec:res}

\subsection{Quantitative results}

\begin{figure*}[t]
  \centering
  
   \includegraphics[width=0.8\textwidth]{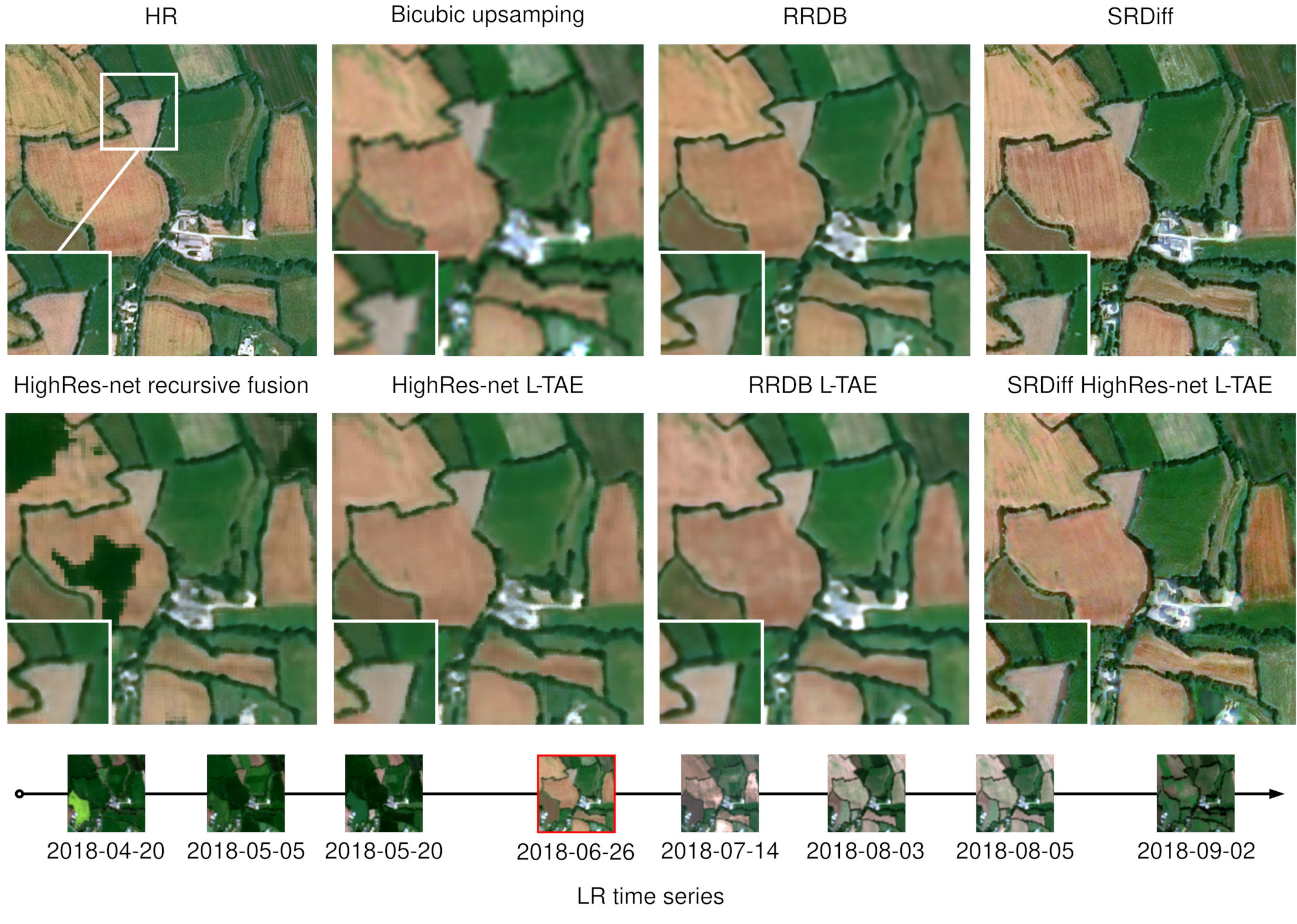}

   \caption{Visual results with the different models. \textbf{SISR:} RRDB, SRDiff (RRDB), \textbf{MISR:} HighRes-net recursive fusion, HighRes-net L-TAE, RRDB L-TAE, SRDiff HighRes-net L-TAE. HR acquisition date: 2018-06-22.}
   \label{fig:visual_results}
\end{figure*}

We report quantitative evaluations of our models in \cref{tab:encoders}, both in SISR and MISR settings.
First, we observe that MAE and Shift-MAE follow the same trends, pointing to a strong sub-pixel co-registration of the two sensors. Then, we observe that SRDiff produces images with the best LPIPS values by a large margin ($\sim 0.3$ compared to $0.6$ for the RRDB backbone alone). This indicates a higher perceptual quality of the super-resolved images. However, SRDiff also obtains worse pixel-based metrics compared to the other models. Conversely, while the other methods have better pixel-based metrics, they present poorer LPIPS results, indicating lower perceptual quality. These observations regarding LPIPS values are indeed corroborated by the qualitative results from \cref{fig:visual_results}. This observation raises an interrogation: should super-resolution strive for perceptual improvements or pixel-wise fidelity? In other applications, such as photography or video processing, perceptual losses, including LPIPS, are a popular choice as the focus is more on the visual quality~\cite{Jo_2020_CVPR_Workshops}. In remote sensing, balancing visual quality and pixel accuracy is crucial for some tasks, requiring further investigation into combining perceptual and pixel-wise losses.

Second, we observe that MISR models tend to outperform their SISR counterparts. For example, RRDB L-TAE outperforms the classical RRDB on all metrics, both reconstruction-based and perceptual ones.
Another notable conclusion is that the temporal attention mechanism added to the time series through L-TAE achieves the best pixel-based metrics values among the methods. In particular, HighRes-net L-TAE outperforms the recursive fusion baseline for all reported metrics, demonstrating the importance of our time-equivariant design. 

Although using stronger conditioning with RRDB enhances the results of SRDiff when it is conditioned simply by the upsampled LR image, SRDiff conditioned by HighRes-net L-TAE produces poorer metrics. MISR models are inherently more computationally expensive and challenging to optimize due to their design, as they process multiple images at once. Their combination with DDPMs, which are also resource-intensive, makes it difficult to achieve optimal results. Therefore, the cost of MISR models is also a concern that must be addressed in future work.

\begin{figure}[t]
  \centering
  
   \includegraphics[width=\linewidth]{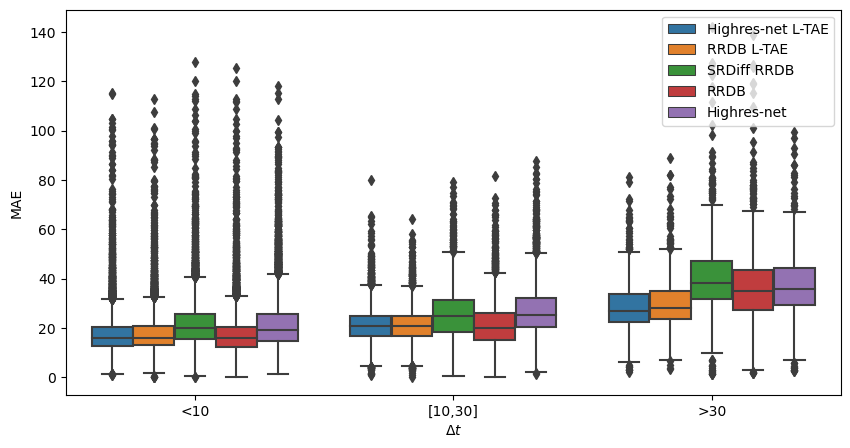}

   \caption{Boxplots of MAE results as a function of the time difference between the acquisitions of the SPOT-6 image and the closest Sentinel-2 image. From left to right: time difference of less than 10 days (10\,758 images), between 10 and 30 days (3\,633 images), over 30 days (630 images).}
   \label{fig:misr_sisr_delta}
\end{figure}

\textbf{Temporal irregularity} To further assess whether L-TAE effectively captures temporal information, we conduct in \cref{fig:misr_sisr_delta} a performance analysis where we compare the MAE results of each model depending on the time difference between the acquisitions of the SPOT-6 image and the closest Sentinel-2 image. Although the dataset has irregular temporal sampling, we observe that time differences between SPOT-6 acquisitions and the closest Sentinel-2 image in the test set are predominantly within 10 days (10\,758 images), with \num{3633} images between 10-30 days, and 630 images above 30 days. This explains why the average scores of SISR models are similar to MISR models. Still, when evaluating the subset of images where the time difference is $>$30 days, MISR models with L-TAE demonstrate lower MAE values than SISR models. It indicates that L-TAE is indeed capturing temporal information in the time series, contrary to HighRes-net, which performs worst.

\textbf{Length of the series} As SITS in real settings can have various lengths, we evaluate our model using different numbers of images. We report in \cref{tab:inputstimeseries} the super-resolution metrics of HighRes-net L-TAE trained for 8 images and evaluated on time series of length $T=\{2, 4, 8\}$, by dropping the images furthest away from $t_\text{HR}$. Intuitively, we observe that more images in the time series result in better reconstruction and perceptual metrics. Yet, we also observe that the drop in performance is not that steep: even with only 2 images, our model outperforms the HighRes-net recursive fusion trained on 8~LR images (\cref{tab:encoders}). This shows that our careful handling of the temporal dimensions helps the network select the relevant images, making it less dependent on additional images that are temporally far from the reference point.

\begin{figure}[t]
  \centering
  
   \includegraphics[width=\linewidth]{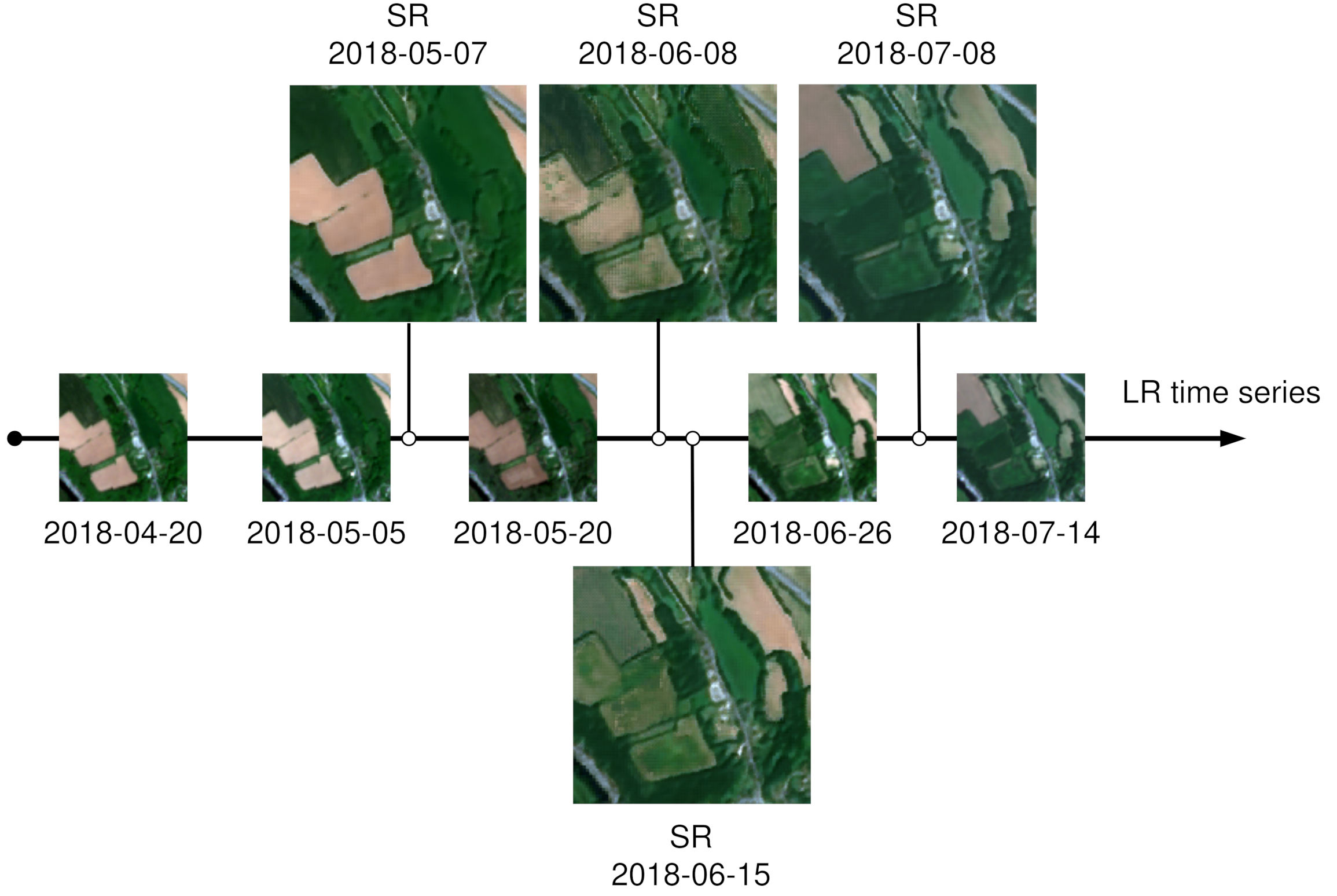}

   \caption{Multi-image super-resolution at different time steps using HighRes-net L-TAE. SR images follow changes in the LR series.}
   \label{fig:change_dates}
\end{figure}

\textbf{Time-equivariant results}
As detailed in \cref{subsec:time_sr}, our method can produce HR images at different times. We show this ability in \cref{fig:change_dates}, with various super-resolved outputs at different timestamps in the series. We observe that the super-resolved images follow the dynamics of the time series, with the field getting greener over time. This illustrates the flexibility of our time-equivariant positional encoding, which can help end-users produce different super-resolved images according to their needs.

\begin{table}
\centering
\resizebox{\columnwidth}{!}{
  \begin{tabular}{@{}lccccccc@{}}
    \toprule
    $N$ & $\downarrow$MAE  & $\downarrow$Shift-MAE & $\downarrow$RMSE    & $\downarrow$LPIPS & $\uparrow$PSNR  & $\uparrow$SSIM \\
    \midrule
8             & \textbf{18.79} & \textbf{18.00}         & \textbf{30.98} & 0.49          & \textbf{20.00}          & \textbf{0.45}\\
4             &19.19 & 18.31     & 31.36 &  0.49 & 19.88 & 0.44 \\
2             &19.95      &18.98           &32.42       &0.49       &19.59      & 0.43      \\
    \bottomrule
  \end{tabular}
}
  \caption{Influence of the length of the time series ($N$) at the inference time. Results are displayed for HighRes-net L-TAE trained with 8 images.}
  \label{tab:inputstimeseries}
\end{table}

\subsection{Cloud impact}

\begin{figure}[t]
  \centering
   \includegraphics[width=\linewidth]{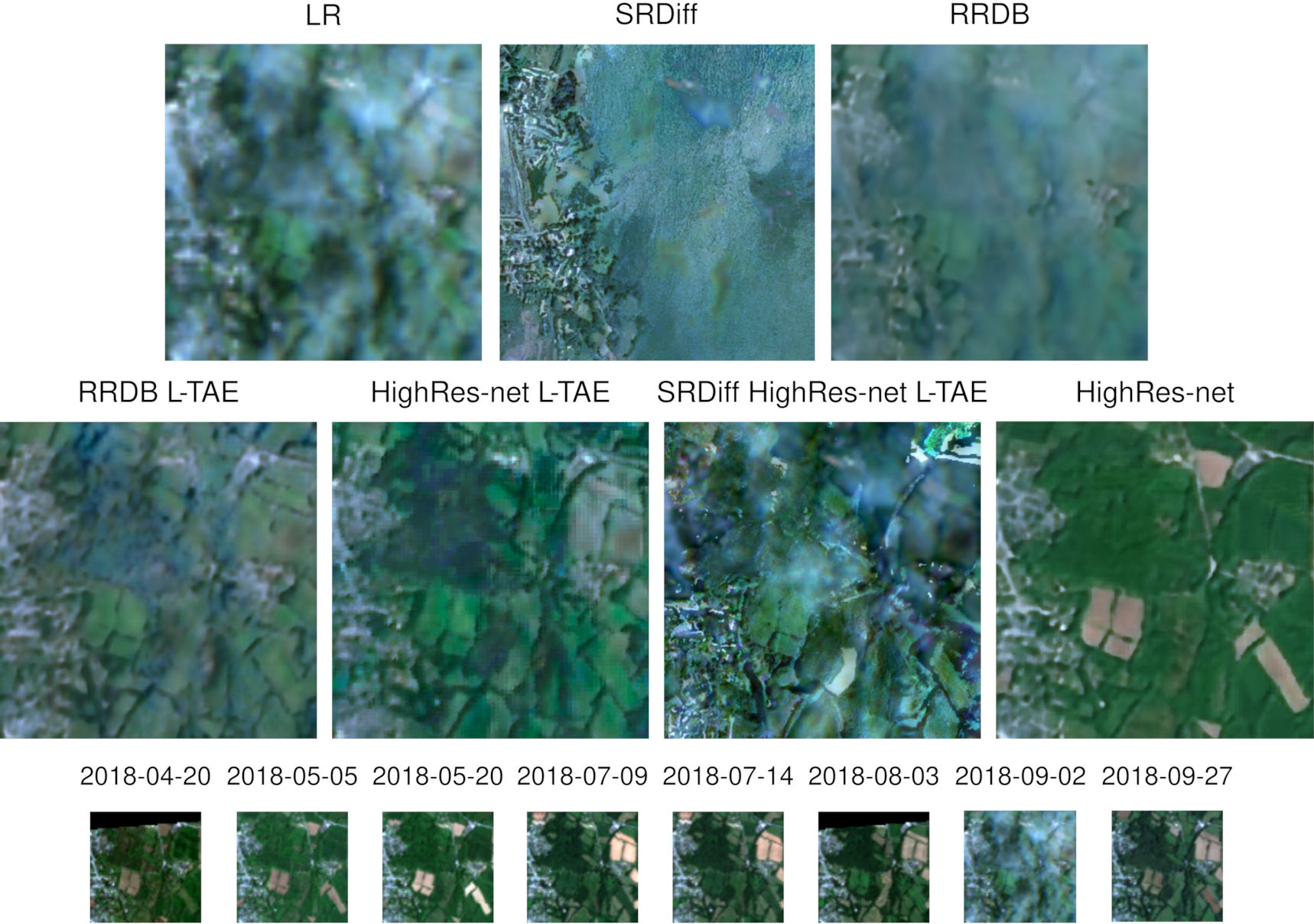}
   \caption{Visual comparison of the effect of clouds. \textbf{SISR:} RRDB, SRDiff (RRDB), \textbf{MISR:} RRDB L-TAE, HighRes-net L-TAE, SRDiff HighRes-net L-TAE, HighRes-net recursive fusion.}
   \label{clouds}
\end{figure}

To minimize the impact of clouds obscuring the landscape and making super-resolution predictions challenging, we gathered Sentinel-2 images with a maximum cloud cover of 5\%. While this criterion significantly reduced the presence of clouds, some images may still contain cloud cover. As shown in \cref{clouds}, MISR models can effectively generate images with reduced cloud cover compared to the input Sentinel-2 image, even in cases where the latter is heavily obscured by clouds.

Among the MISR models, HighRes-net recursive fusion provides the best visual result, with a completely cloud-free output. However, when comparing the outputs of the MISR models to the corresponding Sentinel-2 time series, the output of HighRes-net L-TAE is more coherent with the time series. In the original image, the crop in the central region is expected to be green, but the HighRes-net model with recursive fusion produces an output with a brown crop. In other words, while HighRes-net recursive fusion produces the most visually pleasing image, HighRes-net L-TAE produces an image that is more faithful to the actual changes in the landscape. This shows that L-TAE is effectively assigning more weight to images close to the date on which we choose to apply the SR. Compared to the other MISR models, the super-resolved image generated by SRDiff HighRes-net with L-TAE fusion contains the highest amount of cloud cover. However, it still demonstrates some improvement in cloud reduction compared to the output of SRDiff RRDB. Even though the images produced by our proposed MISR methods are not completely cloud-free, the results are promising and indicate that the MISR approach is a viable solution.

\section{Conclusion}

In this work, we proposed a time-equivariant formulation for MISR to handle temporally irregular SITS. Our approach takes advantage of the acquisition dates to generate coherent super-resolution images at any specified date and outperforms SISR techniques over our dataset \datasetname. This flexibility can be useful in practice for downstream tasks. This work paves the way for cross-sensor MISR research but also raises several questions. We hypothesize that the use of joint reconstruction and perceptual losses as well as enhanced diffusion models should improve our super-resolution models. Besides, we believe that the super-resolution of the full spectrum offered by Sentinel-2, beyond RGB, deserves further investigation.

{\small 
\paragraph{Acknowledgements}
We thank the support of GDR IASIS for funding this work under the SESURE project, the DINAMIS consortium, CNES/Airbus and IGN for access to the SPOT-6 data, and ESA for access to Sentinel-2 data. During the conduct of this research, Simon Donike received a European scholarship to engage in Master Copernicus in Digital Earth, Erasmus Mundus Joint Master Degree (EMJMD). We thank Dirk Tiede (Uni. Salzburg) for his help and feedback on BreizhSR. This work was performed using HPC resources from GENCI–IDRIS (grant 2022-AD011013003).}